# Fusion of medical imaging and electronic health records with attention and multi-head machanisms

Cheng Jiang, Yihao Chen, Jianbo Chang, Ming Feng, Renzhi Wang and Jianhua Yao

*Abstract*— Doctors often make diagnostic decisions based on patient's image scans, such as magnetic resonance imaging (MRI), and patient's electronic health records (EHR) such as age, gender, blood pressure and so on. Despite a lot of automatic methods have been proposed for either image or text analysis in computer vision or natural language research areas, much fewer studies have been developed for the fusion of medical image and EHR data for medical problems. Among existing early or intermediate fusion methods, concatenation of features from both modalities is still a mainstream. For a better exploiting of image and EHR data, we propose a multi-modal attention module which use EHR data to help the selection of important regions during image feature extraction process conducted by traditional CNN. Moreover, we propose to incorporate multi-head machnism to gated multimodal unit (GMU) to make it able to parallelly fuse image and EHR features in different subspaces. With the help of the two modules, existing CNN architecture can be enhanced using both modalities. Experiments on predicting Glasgow outcome scale (GOS) of intracerebral hemorrhage patients and classifying Alzheimer's Disease showed the proposed method can automatically focus on task-related areas and achieve better results by making better use of image and EHR features.

## I. INTRODUCTION

Deep learning based methods have been popular in medical image analysis tasks, including classification, detection, segmentation, and so on [1]. Compared with natural image datasets such as ImageNet and COCO, medical image datasets are always small, and suffer from inaccurate annotations and unbalanced classes, which often cause trouble such as overfitting for deep learning based methods. To alleviate this, incorporating more information such as electronic health records (EHR) can be a good solution [2].

To fuse the information from image data and EHR data, many pioneer works have been proposed [3]. They are usually classified as early fusion [4]–[6], intermediate fusion [7]–[9] and late fusion [10], [11]. For early fusion, the features of different modalities are first extracted independently and then combined before feeding into a classifier. Intermediate fusion is similar to early fusion in using a single classifier to process combined features, but one modality can affect the other's feature extracting process. Late fusion gets the result of each modality separately, and then fuses the results, by averaging, majority voting or meta-classifier. All the three methods are reported to achieve better results than using any modality data alone.

Early fusion is easier for the incorporating of soft-ware extracted and hand designed features, but the image features need to be transformed to one-dimension to cope with the EHR data. Late fusion can simultaneously benefit from state-of-the-art models for each modality, but the simple algorithm at the decision level does not guarantee a fully combination of the complementary information. Compared with the aforementioned two fusion strategies, intermediate fusion is believed to have potential for better results because it iteratively updates its feature representations to better complement each modality through simultaneous propagation of the loss to all feature extracting models [3]. Despite this, for most intermediate fusion methods, the fusion algorithm often happens between the global spatial pooling layer and the fully-connected layer, due to the different spatial dimensions for image and EHR data. This may cause inadequate fusion of image and EHR data since the dimensionality of the image features has been reduced to 1 before fusion.

The goal of the proposed method is to address the aforementioned problems and combine the complementary information from image and EHR data as much as possible. To achieve this, we propose two strategies in a collaborative approach. First, we propose a spatial attention module that fuses both the information from image and EHR data, and generates an image-grid weight map to multiply the features during the image feature extracting phase. This idea mimics the diagnostic pattern of the doctors, since they often tend to focus on some important locations of the image after reading the corresponding EHR. In addition, we also propose to introduce multi-head mechanism to gated multimodal unit for combining different modality features before the fully-connected layer, this operation is also important since some information can only be well fused at high-levels, and the multi-head mechanism can guarantee a better fusion of different aspect of the information by combining them in different subspaces.

In summary, the contribution of this paper is two-fold: First, we are one of the few works to fuse the information of image and EHR data to weight the visual features in the middle layer of convolutional neural network (CNN). Second, the gated multimodal unit (GMU) which is used in an early fusion way is extended with multi-head machnism and used in an intermediate fusion framework for better efficacy. We

C. Jiang and J. Yao are with the Tencent AI Lab, Shenzhen, China. (corresponding author: J. Yao, e-mail: jianhuayao@tencent.com)
Y. Chen, J. Chang, M. Feng and R. Wang are with the Neurosurgery, Peking Union Medical College Hospital, Peking Union Medical College, Chinese Academy of Medical Sciences, Beijing, China demonstrate the performance of the proposed method with experiments using a self-collected Glasgow outcome scale (GOS) dataset and a public Alzheimer's Disease Neuroimaging Initiative (ANDI) dataset.

## II. RELATED WORK

### A. Intermediate fusion for medical images

Similar to early fusion methods, most intermediate fusion methods also combines image features and EHR by concatenation [7], [12], [13]. This is simple but the effectiveness may vary for different datasets because image and EHR features lie in different space and the simply use of fully connected layer may not be powerful enough to combine different significant features in different scales. This point is validated by [13], in which replication and scale operations are performed on EHR data before concatenation, and the choice of them are experimented by grid search. Inspired by [9], the recent work [14] makes the combination of CNN extracted patch features in different scales by considering mean patch features and EHR features, and the final image features from all the scales and EHR features are concatenated before the final fully connected layer. Wang et al [8] use Long short-term memory (LSTM) to process the EHR data and simultaneously produce a saliency weighted global average pooling operator to replace the classical global average pooling operation for the CNN. For the pooled image features and LSTM extracted clinical features, they also concatenate them and use the fully connected layer to make the prediction. To sum up, above methods all use concatenation for the combination of multimodal features. Despite the classical GMU method [15] is proved to be better than simple concatenation, it is only reported in an early fusion way and applied to movie poster classification problem.

### B. Attention mechanism for medical images

Attention mechanism is well known for its ability to emphasis on import information and suppress irrelevant counter-parts. With the successful application of attention mechanism in natural language processing [16] and natural image processing [17]–[20], attention mechanism has also been popular in medical image processing including classification [21]–[23], segmentation [23]–[26], and medical report generation [27]. Trainable attention can be classified as soft-attention and hard-attention [28]. Compared with hard-attention, soft-attention is more popular since it is differentiable and can be easily learned with back-propogation. The attention mechanism can be applied to the spatial dimension [17], [22], [23], the channel dimension of the features [18] or both of them [20], [21], [24]–[26]. Despite the increasing number of works exploiting attention mechanism in the network for medical image, they seldom combine image and EHR features for attention. Even though the pioneer work [8] which is metioned before combines image and clinical features for attention, the attention mechanism is only restricted to the weighted global average pooling stage.

## III. METHODS

### A. Overview

In this section, we introduce the proposed method. To make a fully combination of image and EHR information, two modules named multi-modal attention and multi-head GMU are proposed. The overview of the proposed framework is depicted in Fig. 1. The network can be trained in an end-to-end manner, with paired image and EHR data as input and the prediction result as output. The EHR information has been used in both multi-modal attention and multi-head GMU modules, it contributes in different ways. As for multi-modal attention, its role is to assist the selection of important visual regions. As for multi-head GMU modules, its information is equally combined with dimension reduced visual information for the final classification. We use 3D Resnet-34 [29], [30] as our backbone network. In fact, deeper network such as Resnet-50, Resnet-101, Resnet-152, Densenet-121 [31], Densenet-169 and so on also work with our framework by our experience. We use 3D Resnet-34 mainly for effiency and we believe many other network arichitectures can also work well with our pluggable modules. Although the multi-modal attention module design is generic, not depending on the image and EHR feature size, the optimal location and number of modules may be different for each application and each backbone network. For the datasets we used in this paper, we empirically found adding multi-modal attention module to relatively high-level features is beneficial, the similar idea is also reported in the previous research [9], [23].

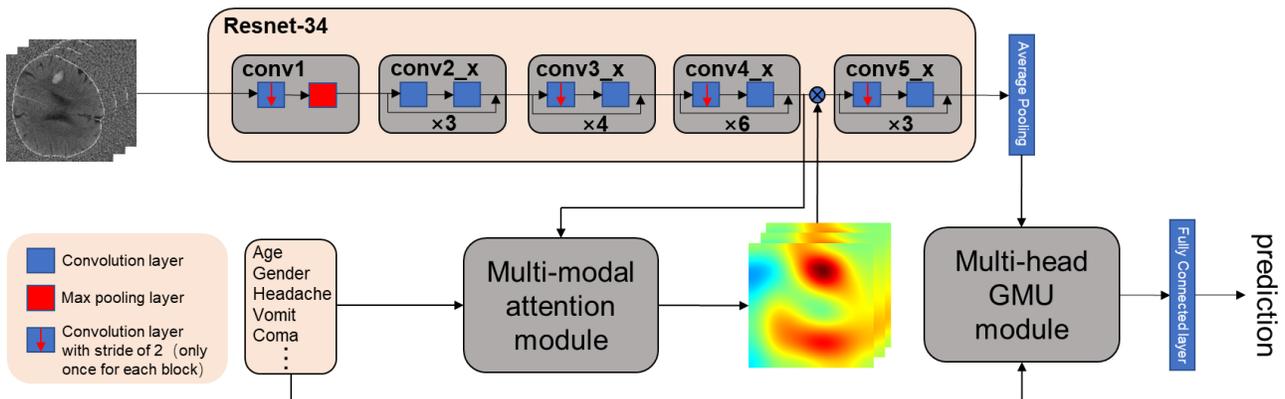

Fig.1 The flowchart of the proposed method.

## B. Multi-modal attention module

Here, we introduce our spatial attention module which can emphasize on salient regions and supresss unrelevant regions through a learned weighting matrix. The proposed attention module belongs to additive attention and is inspired by the work in [17], [23]. The flowchart of the proposed module is shown in Fig. 2.

Since the multi-modal attention module can be incorparted in any existing CNN architecture, we denote $I \in R^{C_I \times H \times W \times D}$ the activation map of a given layer, where $C_I$, $H$, $W$, and $D$ represent the number of channel, height, width, and depth, respectively. The objective is to compute a spatial weight matrix $Q \in R^{1 \times H \times W \times D}$ and output the activation map $I' \in R^{C_I \times H \times W \times D}$ which is computed by multiplying I with Q for every channel.

$$I'_{i,h,w,d} = I_{i,h,w,d} \times Q_{h,w,d} \quad (1)$$

where $i \in 1 \cdots C_I$, $h \in 1 \cdots H$, $w \in 1 \cdots W$, and $d \in 1 \cdots D$.

By tradition, the weight matrix Q is acquired by combination of the local information and the global information. The final feature map vector acquired by global average pooling is usually used as the global information and is seen as task relevant [17]. However, the final feature vector is derived from the front activation map and this combination brings a loop in the network architecture which may cause difficulties in learning the network parameters. For our task, since we have EHR data denoted as $M \in R^{C_M \times 1}$, we can intuitively use it as a task relevant information to replace the role of the global feature vector, and this brings more information to the proposed module and makes the network more compact without any loop.

The weight matrix Q is calculated as follows:

$$F = \sigma_1\big((A_I I + b_I) + Exp(A_M M + b_M)\big) \quad (2)$$

$$Q = \sigma_2(A_F F + b_F) \quad (3)$$

where $A_I \in R^{C_{int} \times C_I}$, $A_M \in R^{C_{int} \times C_M}$, $A_F \in R^{1 \times C_{int}}$ are linear transformations. $b_I \in R^{C_{int}}$, $b_M \in R^{C_{int}}$ and $b_F \in R$ are bias terms. They consist of the parameters of the multi-modal attention module which can be learned during network training. $Exp$ denotes the expansion operation to align matrix dimensions with duplication. $\sigma_1$ is an element-wise nonliearity function and we choose rectified linear unit (RELU) for the proposed module. $\sigma_2$ is the normalization function which make the attention map to be in the range of 0 to 1, and we empirically set it as follows:

$$\sigma_2(Q_i) = (Q_i - Q_{min})/(Q_{max} - Q_{min}) \quad (4)$$

where $Q_i$ is the $i$th element of the attention map, and $Q_{max}$ and $Q_{min}$ are the max and min values of all the elements, respectively.

For the multi-modal attention module, the only hyperparameter is the intermediate dimension $C_{int}$. As suggested in [9], [18], we use $C_{int} = (C_I + C_M)//4$ to limit model compacity and increase the generalization power.

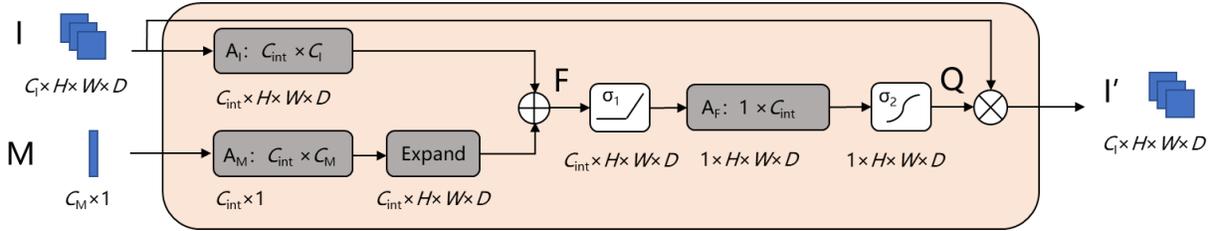

Fig.2 The flowchart of the proposed multi-modal attention module. I and M denote the image features and medical features, respectively. F is the fused features in the hidden space, and Q is the weight matrix for the attention mechanism.

## C. Multi-head Gated Multimodal Units module

Multi-head mechanism allows the model to jointly attend to information from different representation subspaces [16], and GMU is designed to combine multi-modal information in some specific representation subspace [15]. When applied to medical images, one specific representation subspace may not be enough to well fuse the information from the 3D image and EHR data. Therefore, we combine multi-head mechanism and GMU for the fusion of image and EHR features, as shown in Fig. 3. The module consists of multiple GMUs and the output vector is acquired by concatenating all the fused vectors.

Given the visual feature vector $I \in R^{C_I \times 1}$ after global average pooling and EHR feature vector $M \in R^{C_M \times 1}$, we denote the fusion feature vector as $F \in R^{C_{int} \times 1}$, and GMU can be formulated as follows:

$$h_I = \sigma_1(A_I I + b_I) \quad (5)$$

$$h_M = \sigma_1(A_M M + b_M) \quad (6)$$

$$z = \sigma_2\left(A_F \begin{bmatrix} I \\ M \end{bmatrix} + b_F\right) \quad (7)$$

$$F = z * h_I + (1 - z) * h_M \quad (8)$$

where $A_I \in R^{C_{int} \times C_I}$, $A_M \in R^{C_{int} \times C_M}$, $A_F \in R^{C_{int} \times (C_I + C_M)}$ are linear transformations. $b_I \in R^{C_{int}}$, $b_M \in R^{C_{int}}$ and $b_F \in R^{C_{int}}$ are bias terms. $\sigma_1$ and $\sigma_2$ are element-wise nonlinearity function, and we use tanh and sigmoid for $\sigma_1$ and $\sigma_2$, respectively.

For the multi-head GMU module, the hyperparamters are the number n of head and intermediate dimension $C_{int}$. For the later, we set it with the same strategy as that for multi-modal attention module. As for the number n of head, we empirically set it for different applications which will be illustrated in the experiment section.

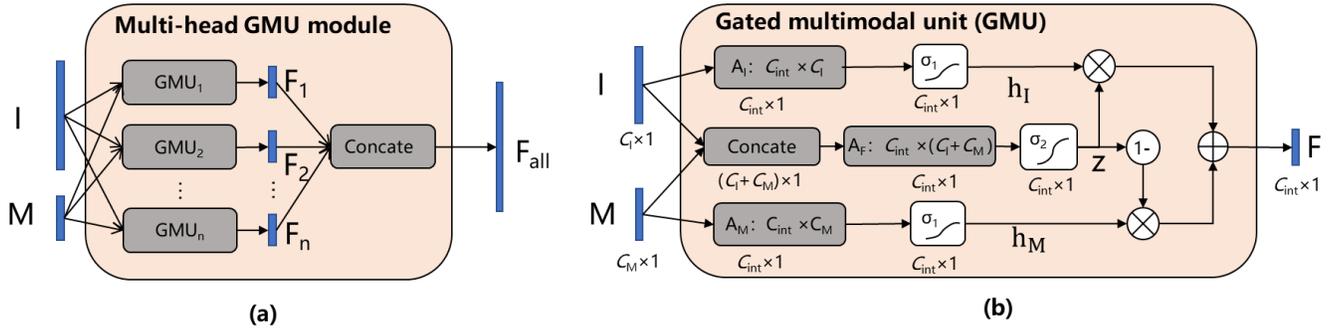

Fig.3 The flowchart of the proposed multi-head GMU module. (a) the flowchart of multi-head GMU.(b) the flowchat of GMU. I and M denote the image features and medical features, respectively. $h_I$ and $h_M$ denote the hidden features of image and medical features, respectively. z is the gate vector, and F is the fused features.

## IV. EXPERIMENTS AND DISCUSSIONS

### A. Evaluation datasets

1) GOS dataset

All data were retrospectively obtained from Chinese Intracranial Hemorrhage Image Database (CICHID). All medical records and CT scans were exported from hospital information system (HIS) and recoded for anonymizing. The characteristics of patients and parameters of scans were collected by independent research assistants. The retrospective study was approved by Institutional Review Board of Peking Union Medical College Hospital (Ethics code:S-K1175).

The data set includes 2486 patients each with one non-contrast computed tomography image (NCCT) and corresponding GOS [32]. The Glasgow Outcome Score applies to patients with brain damage allowing the objective assessment of their recovery in five categories (1:Death, 2: Persistent vegetative state, 3: Severe disability, 4: Moderate disability, 5: Low disability). For simplicity, we category GOS into 2 classes, i.e. poor neurological outcomes ( GOS $\leq$ 3 ) and good neurological outcomes (GOS $>$ 3). We have 1557 patients with poor neurological outcomes and 929 patients with good neurological outcomes. For each patient, we also have corresponding EHR data which contains a total of 21 features: age, sex, Glasgow Coma Scale (GCS), the symptom of headache, vomit, coma, ICH extension to ventricles, history of hemorrhagic stroke, ischemic stroke, hypertension, diabetes mellitus, hyperlipemia, coronary heart disease, heart failure, arrhythmia, anticoagulant therapy, anti-platelet therapy, smoking, alcohol intake, systolic blood pressure, and diastolic blood pressure.

2) ADNI dataset

Data used in the preparation of this article were obtained from the Alzheimer's Disease Neuroimaging Initiative (ADNI) database (adni.loni.usc.edu). The ADNI was launched in 2003 as a public-private partnership, led by Principal Investigator Michael W. Weiner, MD. The primary goal of ADNI has been to test whether serial magnetic resonance imaging (MRI), positron emission tomography (PET), other biological markers, and clinical and neuropsychological assessment can be combined to measure the progression of mild cognitive impairment (MCI) and early Alzheimer's disease (AD). For up-to-date information, see www.adni-info.org.

For each patient, we use his/her first scanned magnetic resonance imaging (MRI) with the description "MPR; GradWarp; B1 Correction; N3". A patient's diagnosis in the dataset is typically categorized as Alzheimer's disease (AD); mild cognitive impairment (MCI); or cognitively normal (CN). In the database, we collect data from 807 patients (AD:181, MCI:399, CN:227). For each patient, we also have the corresponding EHR data which contains a total of 10 features (demographic data: age, gender, years in education, ethnic and racial categories; biofluids: APOe4 genotyping; behavioural assessment: clinical dementia rating – CDRSB; Alzheimer's disease assessment scale – ADAS13; the episodic memory evaluations in the Rey Auditory Verbal Learning Test – RAVLT_immediate; The Mini–Mental State Examination – MMSE).

### B. Implementation Details

All the codes were implemented using python 3.6.8 and Pytorch 1.5.1. Four NVIDIA Tesla V100 GPUs with 32GB memory were used together for training. For the optimization algorithm, we used Adam with default parameters. For GOS dataset, the input size of CT is $192 \times 192 \times 80$, the batch size we used is 32, and the learning rate we used is $3 \times 10^{-4}$. For ADNI dataset, the input size of MRI is $256 \times 256 \times 170$, the batch size we used is 16, and the learning rate we used is 0.01. For the two datasets, we used a warm-up strategy for the beginning of the training, i.e. we set the learning rate 1/5 of the original one and increase it by 1/5 for the first five epochs. In addition, we used ReduceLROnPlateau for learning rate decay, we set the factor parameter as 0.2 for both the datasets, the patience parameter as 3 for GOS dataset and 5 for ADNI dataset. The maximum epoch was set as 50 for both the datasets. All the input images were normalized to mean 0 and variance 1 before feeding into the network.

### C. Evaluation protocol

We compared the proposed method with the method only exploiting image data, the method only exploiting the EHR data, and 3 different intermediate fusion methods to combine both image and EHR data. For a fair comparison, the network backbone and data prepocessing method were kept the same for all the compared methods:

**3D-CNN**: traditional 3D Resnet-34 that only use the image data.

**Tabnet:** the state-of-the-art method that designed for tabular data leanrning and classification named Tabnet [33]. For Tabnet, only EHR data is used.

**Concatenation**: the most common method for fusion image and EHR data that simply concate the pooled image feature and the EHR feature right before the fully connected layer.

**Linear Sum:** Similar to Concatenation but use a linear transformation for each modality so that both outputs have the same size to be summed up, instead of the concatenation operation.

**GMU**: use the original GMU module (single head) to fuse the information of image and EHR data right before the fully connected layer.

For the two dataset, we both use 5-fold cross validation to compare the performance of different methods. For GOS dataset, classic Areas Under Curve (AUC) is used and for ADNI dataset, classic Overall Accuracy (OA) is used for reflecting the overall performance of different methods.

### D. Classification results

TABLE I
The performance of different methods on GOS and ADNI dataset. Each reported value is the mean value with stand devication for the 5-fold cross-validation, and the values for GOS and ADNI dataset denote AUC and OA, respectively.

|  | GOS dataset | ADNI dataset |
|---|---|---|
| 3D-CNN | 0.8593±0.0122 | 0.4969±0.0058 |
| Tabnet | 0.7961±0.0399 | 0.8316±0.0217 |
| Concatenation | 0.8633±0.0145 | 0.8266±0.0425 |
| Linear Sum | 0.8663±0.0122 | 0.9256±0.0119 |
| GMU | 0.8677±0.0133 | 0.9243±0.0122 |
| Proposed | **0.8749**±0.0117 | **0.9343**±0.0104 |

The performance of different methods on the two datasets is reported in Table I. Compare 3D-CNN with Tabnet, we can see that CT images contain more information than EHR data for GOS dataset. But for ADNI dataset, EHR data contributes the most, and little information can be obtained from MRI images for classification. For three class classification, the result of 3D-CNN for ADNI dataset may be lower than some reported studies using ADNI dataset. This is mainly because of the different strategies to select and split the dataset. We split the dataset by patients, while a number of previous studies [34]–[36] split the dataset by MRI images which means a given patient's scans from different visits could exist across both the training and testing sets [37]. Since the prepocessing step does not affect the analysis of the proposed modules, and the preprocessing step is complex with unknown loss of data [38], we choose to use the downloaded ADNI data directly for all the experiments in this paper.

For Concatenation method, the result is only a little lower than those of Linear Sum and GMU methods for GOS dataset but much lower for ADNI dataset. This reflects the drawback of Concatenation method which is not stable when big difference exists between the two modalities. The performance of Linear Sum and GMU are similar for both datasets. The advantages of GMU over Linear Sum is reported in [15], and we think the similar performance may due to the limit number of patients in our datasets. As for the proposed method, it achieves the best performance for both GOS and ADNI datasets, and the results have been largely improved over any one-modality method which validate the importance of fusing multi-modal data for medical diagnosis.

### E. Evaluation of Multi-modal attention module

We first visualize the attention map as shown in Fig. 4. For every dataset, we random chose 4 patients for display. For GOS dataset, we can see the multi-modal attention module can emphasis on the area around the location of intracerebral hemorrhage, it implies the image features around intracerebral hemorrhage is more relevant to the GOS value. As for ADNI dataset, the multi-modal attention module can focus on the brain area and seems to slightly emphasis on the area around hippocampus.

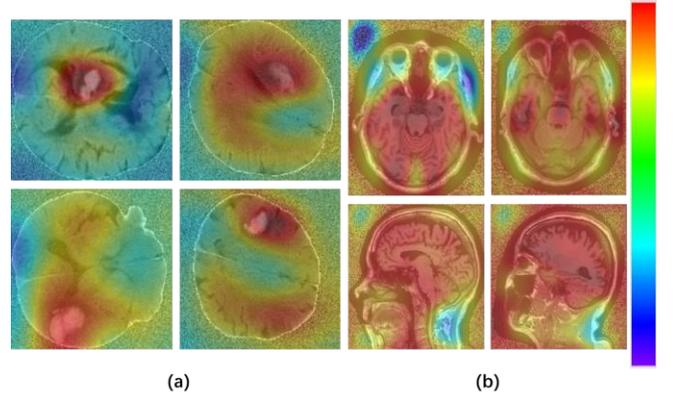

Fig.4 Images overlayed with the attention map. (a) CT images of GOS dataset. (b) MRI images of ADNI dataset. The value of the attention map from low to high is assigned blue to red colors.

To further prove the effectiveness of the proposed multi-modal attention module, we also compare it with other well-known attention module for GOS dataset, i.e. spatial attention in [23] named Grid attention, channel attention in [18] named SE attention, and Non-local attention in [19]. For a fair comparison, all attention modules are used in the same position as the proposed method and we keep the other settings the same. The result is shown in Table II. From Table II, the proposed multi-modal attention module achieve the best performance which imply that by combining the information of EHR data and image data, better attention machnism can be expected.

TABLE II
The performance of different attention modules. Each reported value is the mean value of AUC with stand devication for the 5-fold cross-validation of GOS dataset. For a fair comparison, all attention modules are used in the same position as the proposed method and we keep the other settings the same.

| Grid | SE | Non-local | Proposed |
|---|---|---|---|
| 0.8691 | 0.8712 | 0.8706 | **0.8749** |
| ±0.0149 | ±0.0148 | ±0.0171 | ±0.0117 |

### F. Evaluation of Multi-head Gated Multimodal Units module

To test the effect of multi-head GMU module, results are shown in Fig. 5 with different head numbers for both datasets. As Fig. 5 shows, better results can be expected when the head number is above 1 which validate the necessity of the multi-head mechanism. Moreover, larger head numbers does not mean monotonous growth of the performance, the optimal number may vary for different applications and datasets, and remains as an open question.

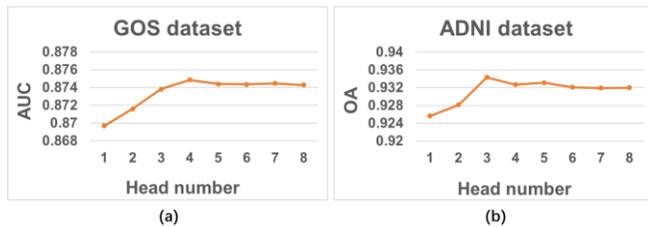

Fig. 5 Performance of the multi-head GMU module with the variation of head numbers.

## V. CONCLUSIONS

We proposed a new deep learning based method in this paper for image and EHR data fusion, to improve the precision of medical diagnosis. The proposed multi-modal attention module combines the information from both image and EHR features to allow the following CNN layers focus on more related regions. The experiments have proved this both visually and quantitatively. Despite image features can be better extracted, the fusion of the final extracted image features and EHR features are also important because it is closely related to the final decisions. In this paper, we further pointed out multi-head mechanism is also suitable for GMU module, making it better than other existing feature combination strategies. Experiments demonstrating the effectiveness of the proposed method were presented, and the proposed method achieves the best for both GOS estimation and Alzheimer classification tasks.

Compared with existing attention modules, we showed that injecting more information from other sources can be a good thought to improve the attention machnism which may also be useful for other applications. In addition, we should also pay attention to the multi-head mechanism which is usually useful and practical to improve the results. Moreover, the proposed two modules do not rely on any specific CNN backbones, which also benefit further applications and extensions.

For the datasets used in this paper, the number of samples are still a stumbling block for the overall performance and analysis. We believe with more samples, not only better performance can be expected, but also help to better ascertain the performance gain obtained by any future enhancements. Further research to use image features to guide the selection of EHR features may also be interesting.

To summarize, the proposed method is useful for clinical research by suggesting related regions, and can also be used to build a computer-assisted diagonosis (CAD) system to help the doctors to make decisions.